\documentclass[10pt,twocolumn,letterpaper]{article}

\usepackage{times}
\usepackage{epsfig}
\usepackage{graphicx}
\usepackage{amsmath}
\usepackage{amssymb}
\usepackage{subcaption}
\usepackage[colorlinks=true]{hyperref}

\date{\vspace{-5ex}}

\begin{document}

\title{Autonomous driving challenge: To Infer the property of a dynamic object based on its motion pattern using recurrent neural network}

\author{Mona Fathollahi and Rangachar Kasturi\\
Department of Computer Science and Engineering\\
University of South Florida, Tampa\\
\tt\small (mona2,r1k)@mail.usf.edu\\
}
 
\maketitle

\begin{abstract}
In autonomous driving applications a critical challenge is to identify the action to take to avoid an obstacle on a collision course. For example, when a heavy object is suddenly encountered it is critical to stop the vehicle or change the lane even if it causes other traffic disruptions. However, there are situations when it is preferable to collide with the object rather than take an action that would result in a much more serious accident than collision with the object. For example, a heavy object which falls from a truck should be avoided whereas a bouncing ball or a soft target such as a foam box need not be.

We present a novel method to discriminate between the motion characteristics of these types of objects based on their physical properties such as bounciness, elasticity, etc.

In this preliminary work, we use recurrent neural network with LSTM (Long Short Term Memory) cells to train a classifier to classify objects based on their motion trajectories. We test the algorithm on synthetic data, and, as a proof of concept, demonstrate its effectiveness on a limited set of real-world data.

\end{abstract}

\section{Introduction}
In recent years, the technology of self-driving cars has made dramatic progress. One of the critical challenges of this emerging technology is the safety of both car occupants and other road users. The current prototype of autonomous cars are equipped with advanced sensors such as ultrasonic, vision, radar and LIDAR. These sensors along with sophisticated data fusion algorithms are able to detect and track obstacles in real-time with very good resolution.  

When an obstacle is detected in the planned path, either its planned route should be modified or the vehicle should come to stop. Depending on the traffic situation and vehicle speed, this policy could cause collision with other vehicles. Therefore, obstacle avoidance may not always be the safest action. Similar challenge has been discussed in \cite{ramos_2015}. 

The intuitive solution would be to recognize the object before taking an action. The intelligent unit should predict whether it is safe to pass over the object or it should inevitably follow avoiding policy.

\begin{figure}[t]
\begin{center}
    \begin{subfigure}{.37\columnwidth}
      \centering
      \includegraphics[width=.95\linewidth]{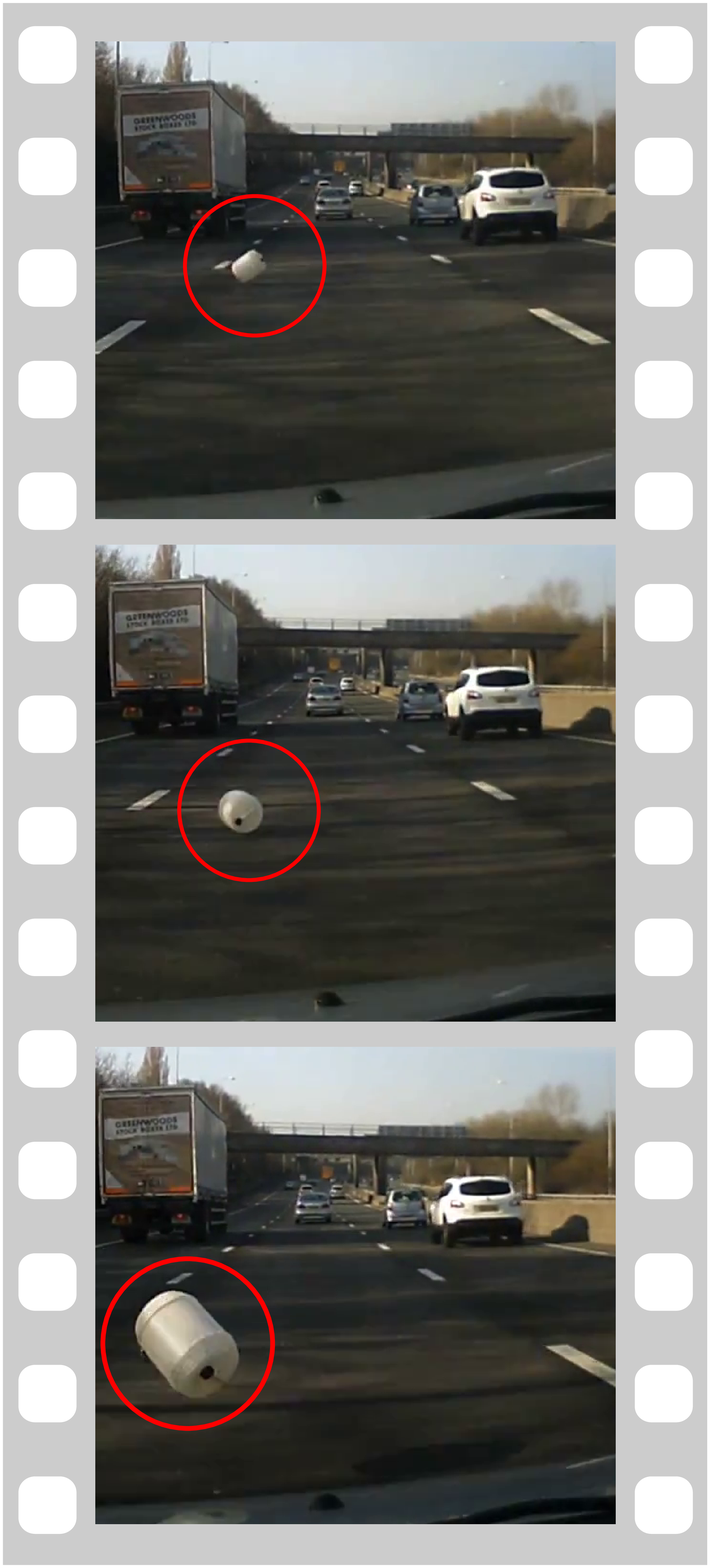}
      \caption{}
      \label{fig:sfig1}
    \end{subfigure}%
    \begin{subfigure}{.37\columnwidth}
      \centering
      \includegraphics[width=.95\linewidth]{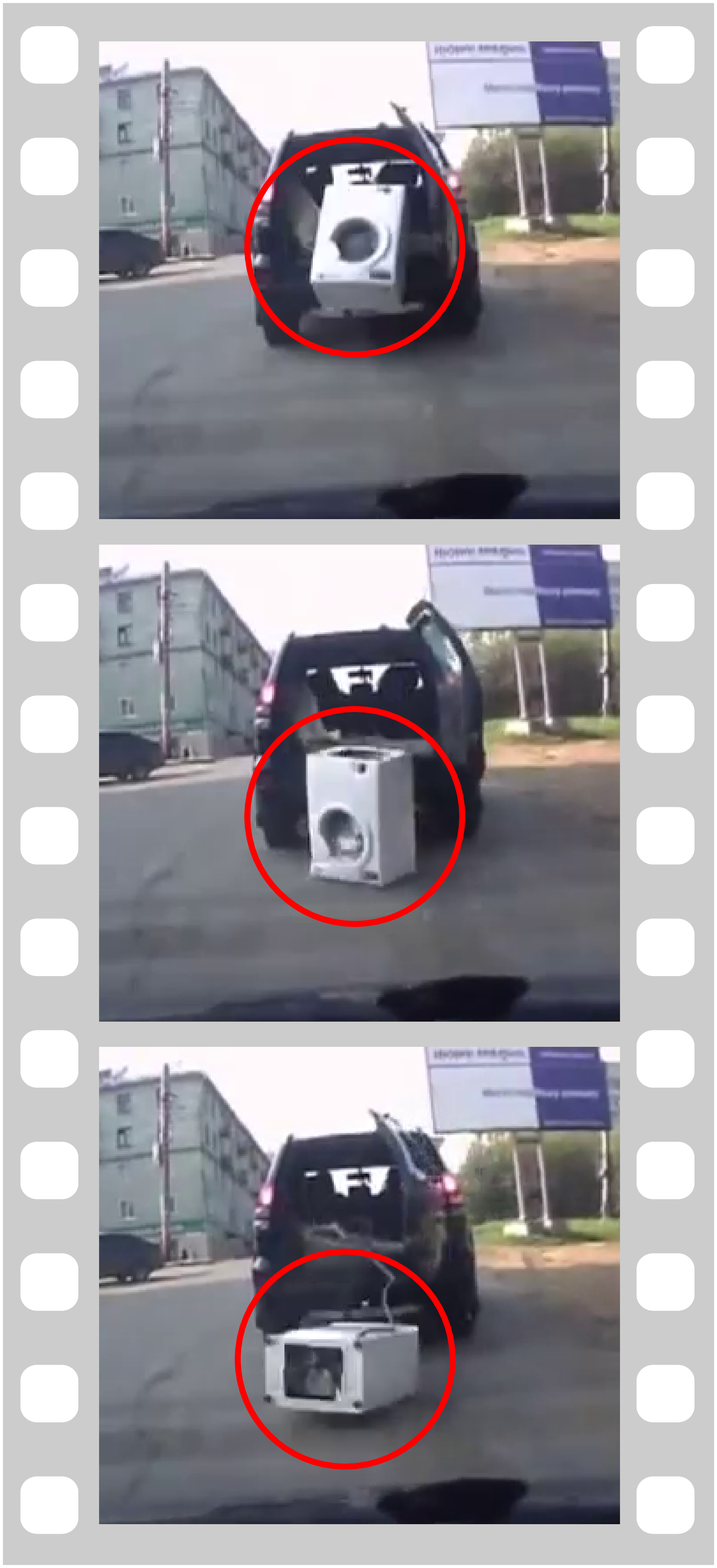}
      \caption{}
      \label{fig:sfig2}
    \end{subfigure}
\end{center}
    \caption{Selected frames of dynamic objects on the road. (a) A plastic container which is safe to collide, \href{https://www.youtube.com/watch?v=bIm-ffb-SKs&feature=youtu.be&t=14}{YouTube Link}. (b) a heavy object that should be avoided, \href{https://www.youtube.com/watch?v=dByja-adlDY&feature=youtu.be&t=24}{YouTube Link}.}
\label{fig:obstacle}
\end{figure}

A sample video for each scenario is downloaded from Youtube and a few frames are shown in the Figure \ref{fig:obstacle}. In the first video, an empty plastic container is bouncing in the road which is safe to pass. In the second video, a heavy object is falling out of the front car which should definitely be avoided.

The immediate solution that one might consider is to formulate the problem as a regular image classification task and collect a dataset of collision safe and unsafe objects. While there is much progress in object detection/recognition methods \cite{lecun2015deep}, this approach has several challenges which makes it ineffective for this particular application. 

First, collecting a dataset that contains different objects in different lighting conditions and viewpoints is a difficult task in itself. Second, it is almost impossible to infer the weight of an object by its visual cue; for example, two very similar boxes with one of them filled with metal pieces and the other one which is empty have similar images. 

Finally, there is a high possibility of recognition failure because the image resolution is usually poor for far away objects. Also, the classifier should decide in a short period of time, where motion blur might make the problem even more challenging. For example, the white plastic container in the first column of Figure \ref{fig:obstacle} could be classified as a gas cylinder. 

These challenges are easily resolved by a human by observing the trajectory of empty box versus heavy box (e.g. plastic container versus a gas cylinder). Therefore, assuming that the real-time trajectory of the dynamic object is available \cite{bewley2014online}, we claim that motion pattern provides strong cue to infer the object dynamics accurately and to classify it as a "safe to pass over" or "must avoid" object.

\section{Method}
In this section, our goal is to design a classifier to infer object's bounciness characteristic based on its trajectory when it hits the ground. 
Our approach is based on the observation that the bouncing pattern of objects is directly affected by their mass. 

\begin{figure}[!t]
\begin{center}
    \includegraphics[width=.65\columnwidth]{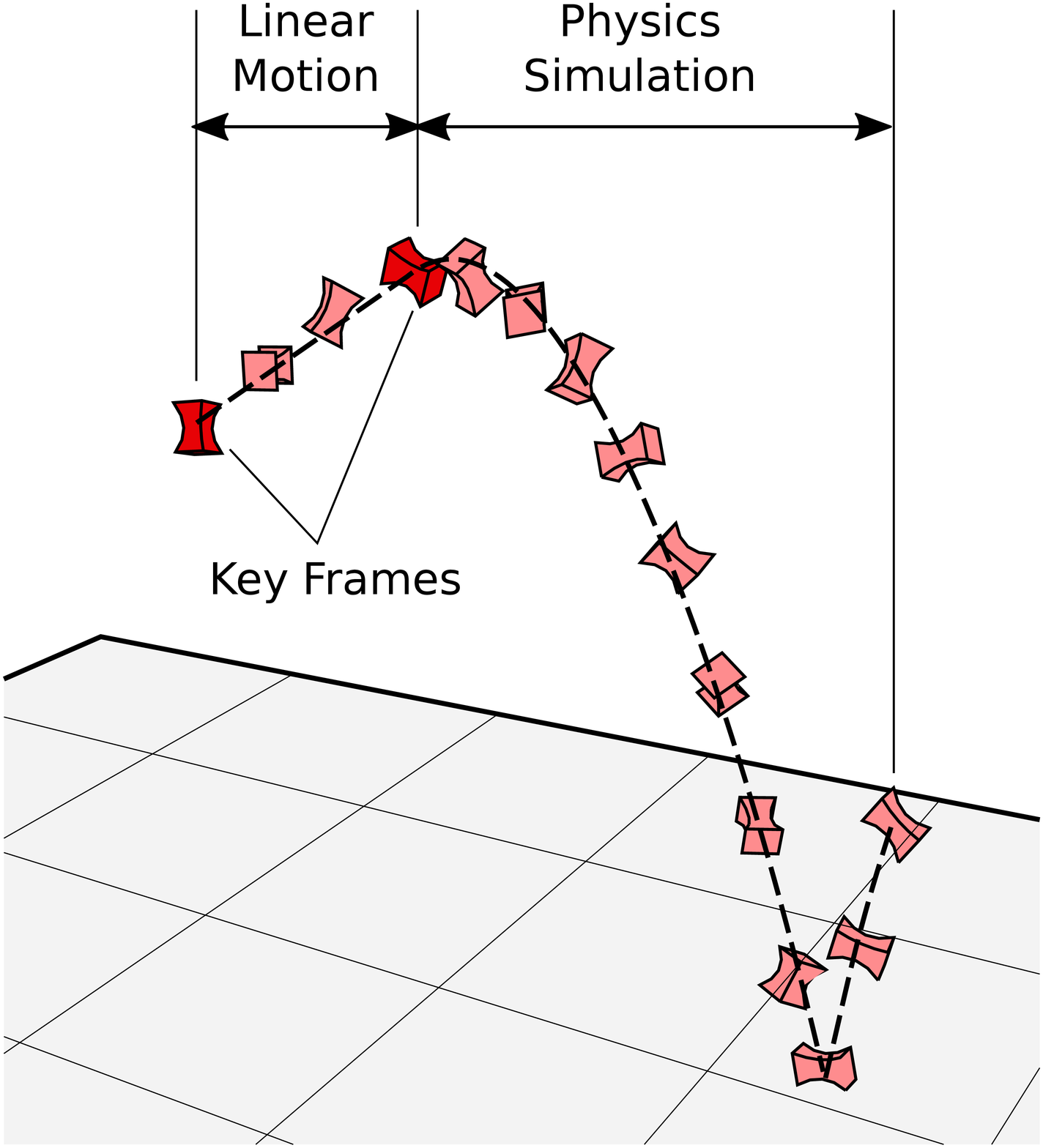}
\end{center}
    \caption{Object trajectory synthesis in Blender. The first few frames are to generate initial velocity (Linear motion). The second part, physics simulation, is recorded as trajectory of the object.}
\label{fig:blender}
\end{figure}
 
\subsection{Data}
To collect data, we should throw different objects with different masses and shapes and record their trajectories. On the other hand, since the bounciness of the object is also related to initial velocity, we should collect a large amount of data to be able to learn the effect of mass on the trajectory. Therefore, dataset collection in this case is cumbersome and expensive. 

Therefore, we generate synthetic videos with binary labels denoting heavy or light object trajectory. We utilized open source 3D creation suite, Blender \cite{blender}, to generate motion data of bouncing objects. Blender uses “Bullet Physics Library” for collision detection, rigid body dynamic simulation and other Physics simulations tasks.

Each trajectory starts from random coordinates and Euler angles and the object has random initial linear and angular velocities. To generate random initial velocities, two key-frames are inserted at first and seventh frames. Also, the height of object at first frame and both linear and angular positions at seventh frame are randomized. (Figure \ref{fig:blender}). The physics engine takes over the object animation after seventh frame. The world coordinates of the object after the seventh frame are recorded as object trajectory time series.

In the initial phase of our project, we only consider two object categories; the first class is the trajectory of light objects that have a high tendency to bounce when they hit the ground, and the second class are the objects that are heavy and have more tendency to slide than bounce. To isolate the effect of shape on the bounciness, we kept the shape of the objects the same for all simulations. 
We generated 1000 training videos, and 1000 test videos for both categories.

\begin{figure*}[!t]
    \centering
    \begin{subfigure}[b]{0.31\textwidth}
        \centering
        \includegraphics[width=\textwidth]{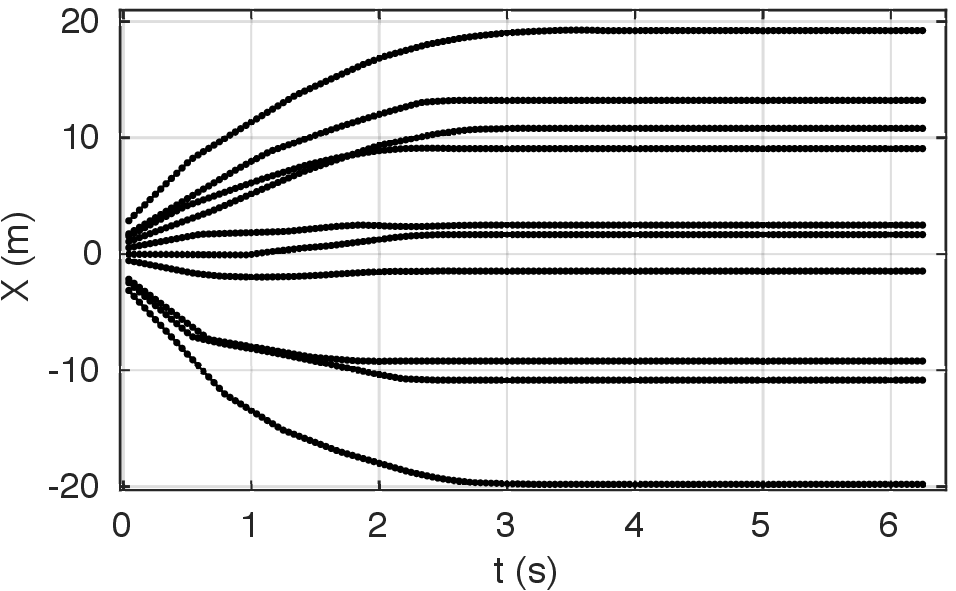}
    \end{subfigure}
    \begin{subfigure}[b]{0.31\textwidth}  
        \centering 
        \includegraphics[width=\textwidth]{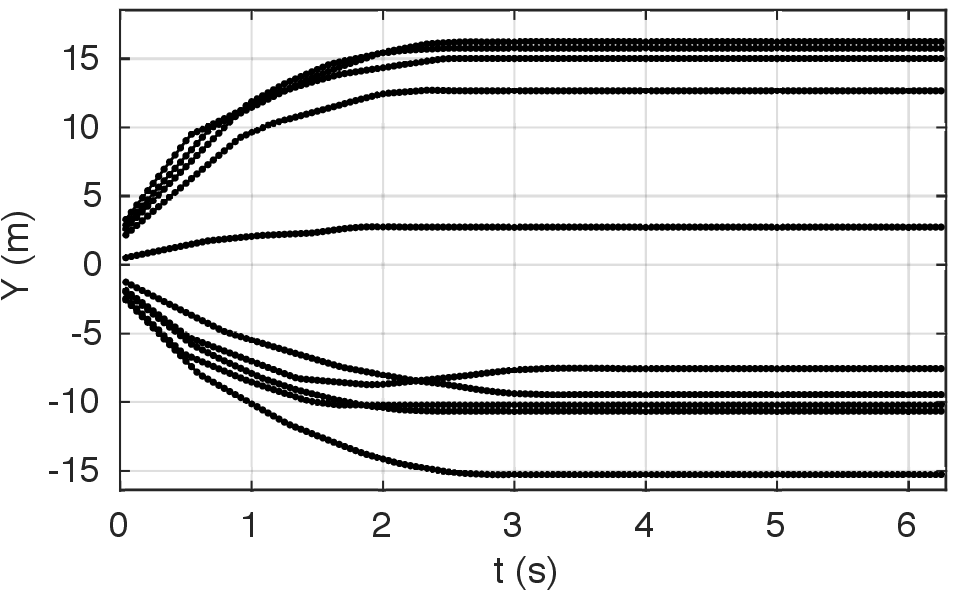}
    \end{subfigure}
    \begin{subfigure}[b]{0.31\textwidth}  
        \centering 
        \includegraphics[width=\textwidth]{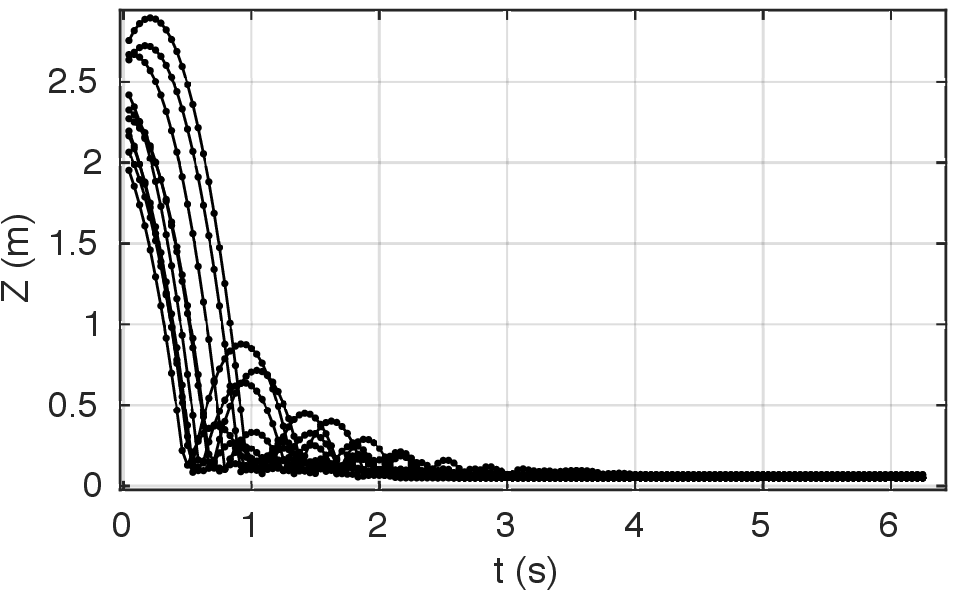}
    \end{subfigure}
    \vskip\baselineskip
    \begin{subfigure}[b]{0.31\textwidth}
        \centering
        \includegraphics[width=\textwidth]{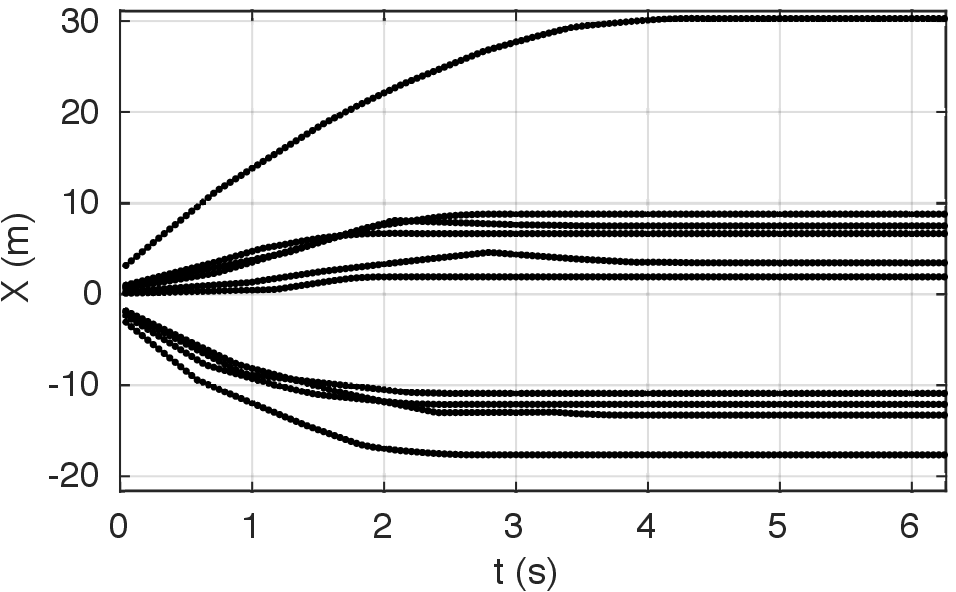}
        \caption{}    
    \end{subfigure}
    \begin{subfigure}[b]{0.31\textwidth}  
        \centering 
        \includegraphics[width=\textwidth]{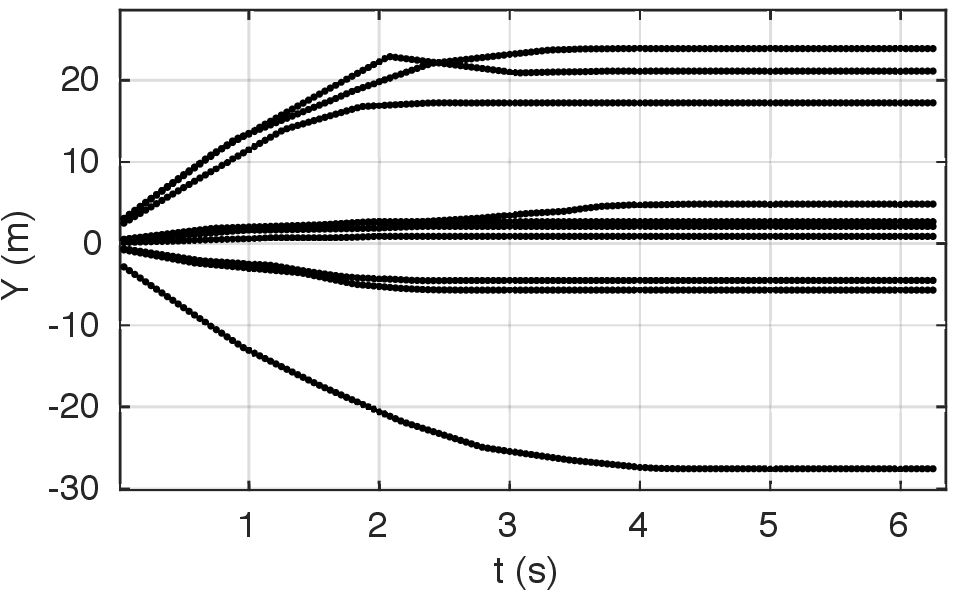}
        \caption{}    
    \end{subfigure}
    \begin{subfigure}[b]{0.31\textwidth}  
        \centering 
        \includegraphics[width=\textwidth]{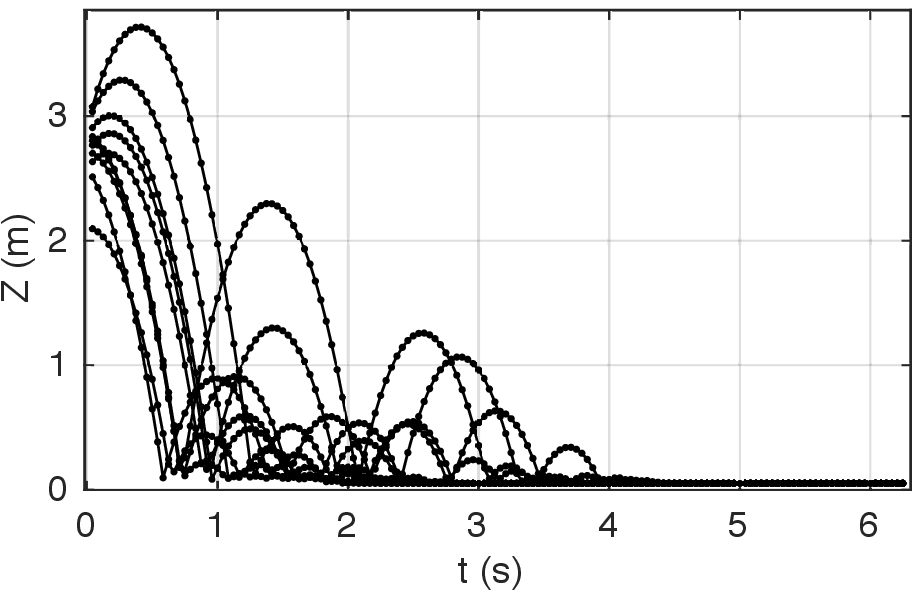}
        \caption{}    
    \end{subfigure}    
    \caption{Samples of synthesized trajectory data set. The figures shows time plots of (a)X, (b)Y, (c)Z coordinate of the (top)heavy, (bottom)light object.}    
    \label{fig:trajs}
\end{figure*}

Some randomly chosen examples of the generated trajectory data are shown in Figure \ref{fig:trajs}. Even though there is a clear distinction between Z dimension of the trajectories, we still see subtle difference in X and Y components. For example, for a light object (higher bounciness) it takes more time to come to a full stop and this is reflected in X and Y coordinates and this justify the superior performance of classification when 3D data is used (Table \ref{tab:bestAcc}). Finally, although some statistical differences are detectable between the two categories, the plots in this figure suggests that no simple rule can be proposed based on, for example, the number of bounces or time series duration; therefore, a more involved classification algorithm is required.

\subsection{Classifier}

In this section, we assume that the trajectory of the object is given by a tracking algorithm. Therefore the problem is reduced to time series classification.

In this work, we adopted, Recurrent Neural Network models (RNN) for trajectory classification. RNN is a type of artificial neural network that is able to process data with arbitrary input sequence lengths. Their internal memory units and feedback loops have made them a very successful tool in sequential data prediction in various domains \cite{donahue2015long} \cite{graves2014towards}. Recently, they have been used in the context of time series classification \cite{shah2016applying}. In this work we also use RNN architecture with peePhole Long Short-Term Memory (LSTM) units \cite{sak2014long} for motion trajectory classification. 

The input to the network is the first T seconds of objects' trajectory. Training and test time series are normalized by their standard deviations in each dimension. Our network architecture is a two-layered LSTM with 64 hidden units in each layer. The hidden state at the last time step of LSTM is fed into a softmax layer. We also add a dropout layer between second LSTM layer and softmax layer with rate 0.8. To compute parameter gradients, the truncated back-propagation-through-time (BPTT) approach \cite{williams1990efficient} is used to reduce the probability of vanishing gradient problem for long input sequences. The entire implementation is done using Tensorflow \cite{abadi2016tensorflow} package.

\begin{table}[!b]
\centering
\caption{Best accuracy for 3D and 1D input sequence.}\label{bestAcc}
\begin{tabular}{ll}
\hline
Trajectory dimension(s) & Best Accuracy(\%) \\ \hline
X, Y, Z               & 81              \\
only Z                & 78      \\   
\hline
\end{tabular}
\label{tab:bestAcc}
\end{table}

 \begin{figure}[!b]
\begin{center}
    \includegraphics[width=0.75\columnwidth]{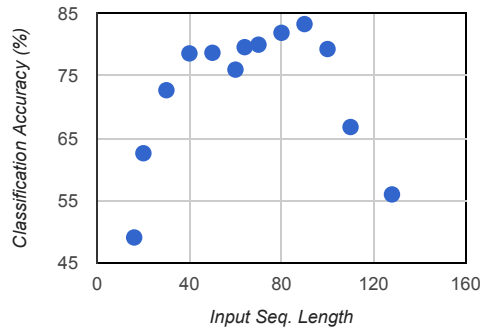}
\end{center}
    \caption{Classification accuracy versus trajectory length.}
\label{fig:excel}
\end{figure}

 In the first experiment, the impact of input dimension (XYZ vs Z) on the classification accuracy is studied. We perform grid search on the parameters to get the maximum accuracy when only z-coordinate of the trajectories is used for training. The same parameters are used to train the network with 3D input, the results are compared in Table \ref{tab:bestAcc}. Superior performance was achieve with 3D inputs, because it takes more time for a light object to stop along X and Y direction.
 
In the second experiment, we study the influence of input sequence length on the accuracy. If it is too short, the classifier has limited data and might not be able to learn distinguishable pattern. On the other hand, increasing the input sequence beyond some limit could cause the gradient of LSTM network to start vanishing or exploding, which consequently leads to the accuracy drop. In the Figure \ref{fig:excel}, we have shown the classification accuracy for different lengths of input sequence.

\begin{figure}[!t]
    \centering
    \begin{subfigure}[b]{0.75\columnwidth}
        \centering
        \includegraphics[width=\textwidth]{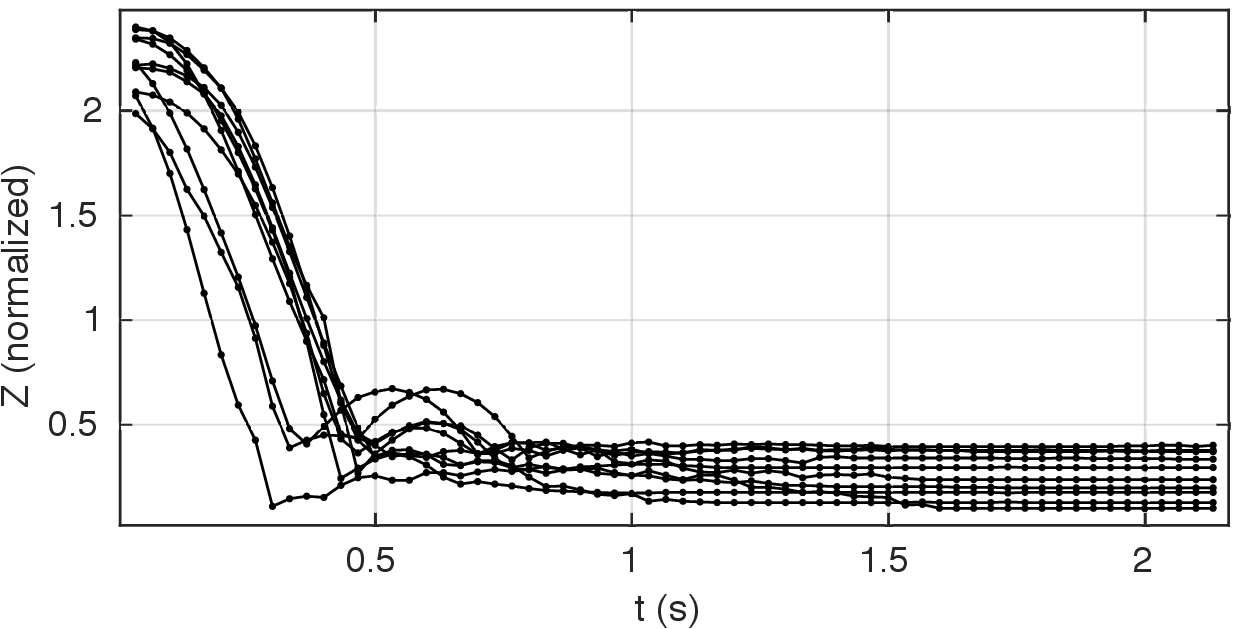}
    \end{subfigure}
    \vskip\baselineskip
    \begin{subfigure}[b]{0.75\columnwidth}   
        \centering 
        \includegraphics[width=\textwidth]{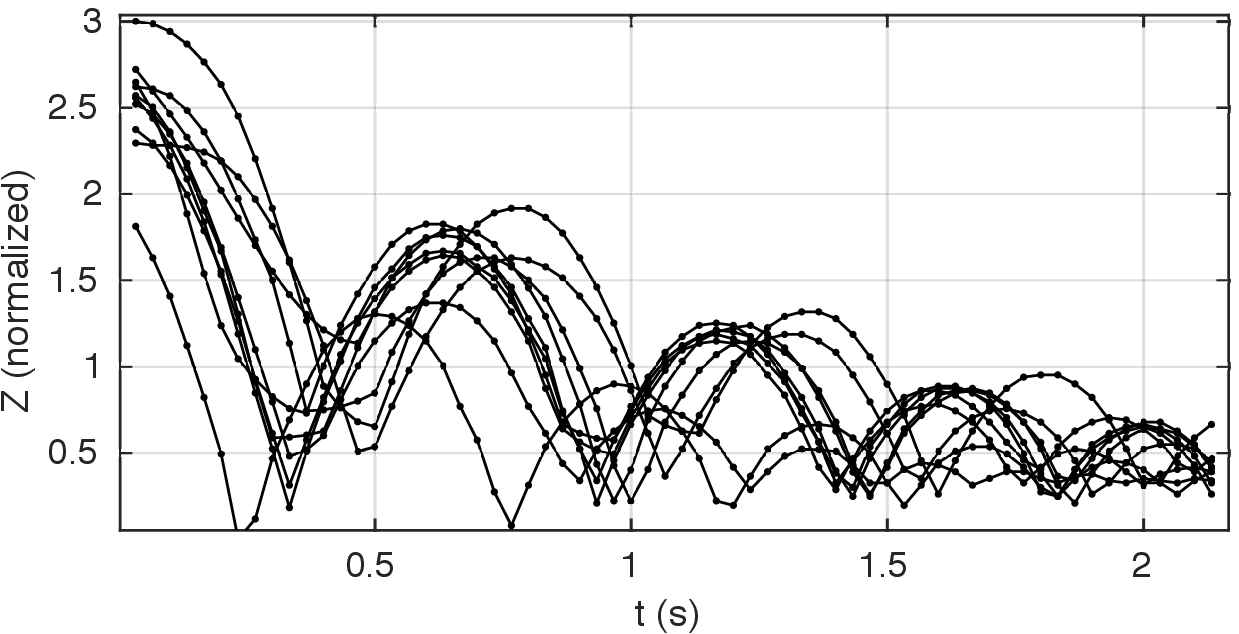}
    \end{subfigure}
    \caption{Real-world trajectories: wooden cube (top), golf ball (bottom).}    
    \label{fig:rtrajs}
\end{figure}

\subsection{Experiment on real-world data}
\begin{figure}[!t]
\begin{center}
    \begin{subfigure}{.5\columnwidth}
      \centering
      \includegraphics[width=0.95\linewidth]{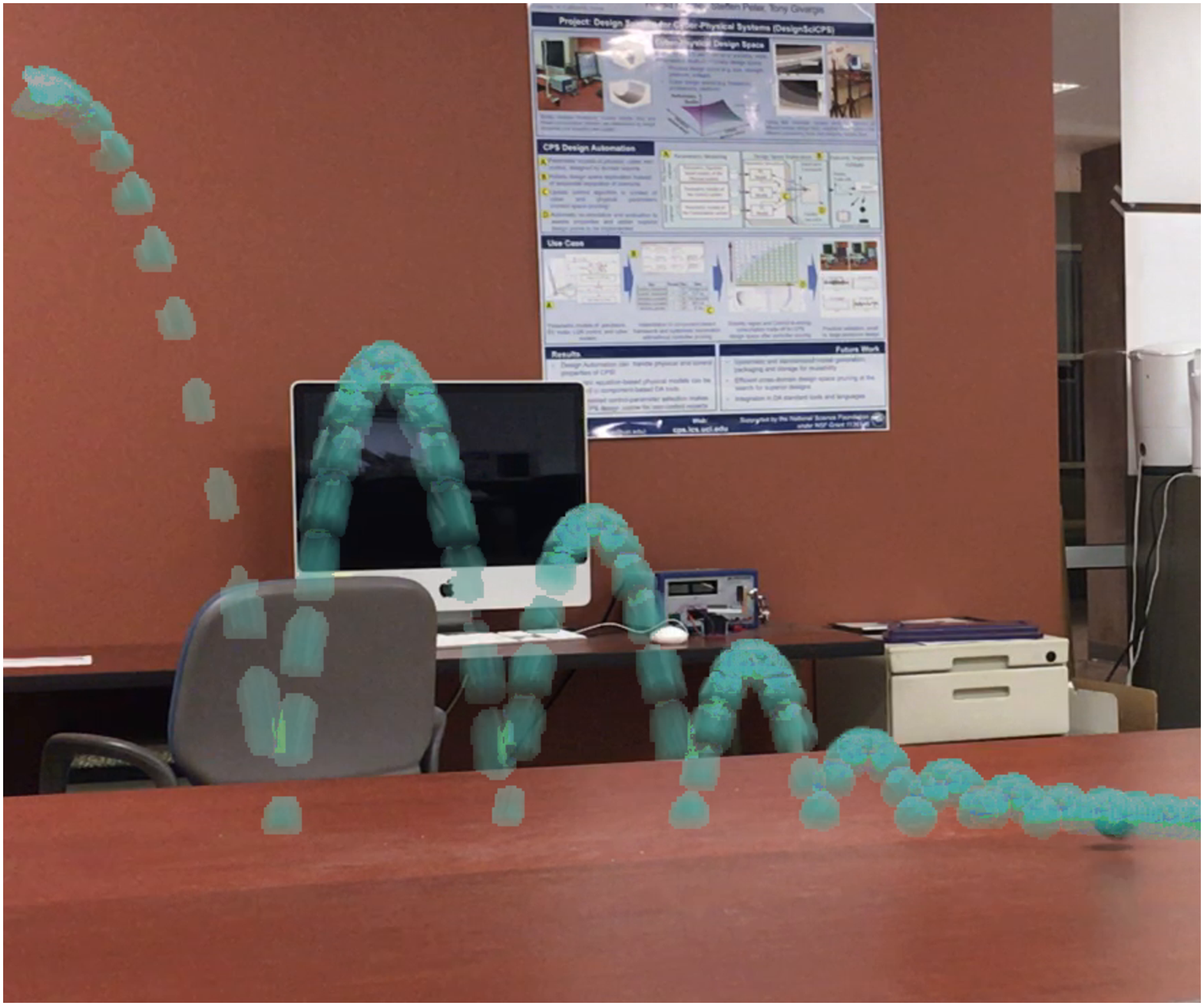}
    \end{subfigure}%
    \begin{subfigure}{.45\columnwidth}
      \centering
      \includegraphics[width=.95\linewidth]{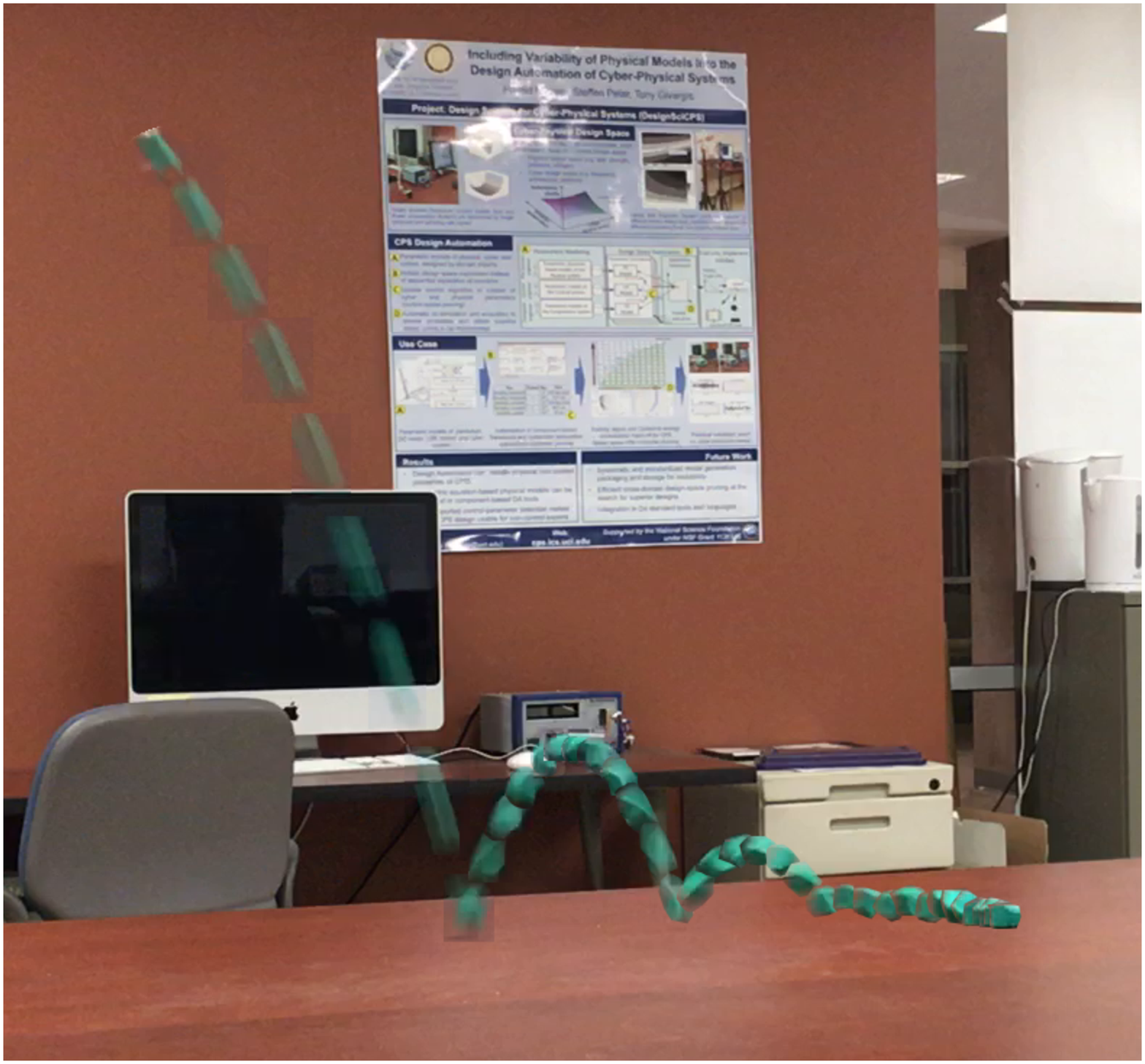}
    \end{subfigure}
\end{center}
    \caption{Real-world experiment sample trajectories: (Left) golf ball (Right) wooden cube.}
\label{fig:real}
\end{figure}

In this section, we leverage the trained network on synthetic trajectories, to analyze the trajectory of real-world objects. One extreme example is chosen from each category: golf ball as an object with high bounciness, and wooden cube as an object with low bounciness. We throw them from different heights with different initial velocities and record the video, Figure \ref{fig:real}. The objects are marked with a distinct color to be able to use a simple color tracker. Lastly, when frames get blurred due to the fast motion of the object, missing part of the trajectory is reconstructed by a simple interpolation.
The trajectories that are shorter than input sequence length get zero-padded. For each category we collected 20 videos and plotted the trajectories in Figure \ref{fig:rtrajs}. In this experiment, the trajectories are recorded with a single RGB camera and only trajectory along z direction is used for decision; therefore we used the trained network on z channel as well. We obtained an accuracy of 93\% on the ball and 100\% on the wooden cube.

 \section{Conclusions and Future work}
In our preliminary experiments, we have found that motion pattern of an object provides strong cue on the object property. This has a potential application in the autonomous driving technology which reduces the number of dangerous stops or maneuvers when on object suddenly appears in front of the vehicle. 
Our preliminary experiments show promising results on synthetic and small set of real-world data. In the future, we are planning to collect more real-world examples to fully develop and test the concepts explored in this paper.

{\small
\bibliographystyle{plain}
\bibliography{egbib}
}

\end{document}